%% file: paper.tex
\documentclass[]{opendatalab}
\usepackage{xcolor}
\usepackage[T1]{fontenc}
\usepackage{longtable}
\usepackage{listings}
\usepackage{upquote}
\usepackage{xspace}
\usepackage[numbers]{natbib}
\definecolor{codegreen}{rgb}{0,0.6,0}
\definecolor{codegray}{rgb}{0.5,0.5,0.5}
\definecolor{codepurple}{rgb}{0.58,0,0.82}
\definecolor{backcolour}{rgb}{0.95,0.95,0.92}
\definecolor{promptcolor}{HTML}{D1D0F2}
\definecolor{promptcolorheader}{HTML}{bdbcec}
\lstdefinestyle{promptstyle}{
  basicstyle=\ttfamily\scriptsize,
  backgroundcolor=\color{promptcolor!50},
  numbers=none,
  breaklines=true,
  breakatwhitespace=false,
  columns=fullflexible,
  keepspaces=true,
  showstringspaces=false,
  upquote=true,
  aboveskip=0.2em,
  belowskip=0.2em
}

\newcommand{\promptheading}[1]{\par\smallskip\noindent\textbf{\#\#\# #1}\par}
\newcommand{\promptcode}[1]{\texttt{#1}}

\newenvironment{promptcriteria}{%
  \begin{enumerate}
  
  \setlength{\itemsep}{0pt}
  \setlength{\parskip}{0pt}
}{%
  \end{enumerate}
}

\newenvironment{promptbox}[1]{%
  \begin{tcolorbox}[
  breakable,
  top=0.3em,bottom=0.3em,left=0.5em,right=0.5em,
  toptitle=0.3em,bottomtitle=0.2em,boxsep=0pt,
  colframe=promptcolorheader,colback=promptcolor!50,boxrule=0.5pt,
  title={#1},
  fonttitle=\footnotesize\bfseries,
  coltitle=white
  ]
  \footnotesize
  \setlength{\parindent}{0pt}
  \setlength{\parskip}{0.25em}
}{%
  \end{tcolorbox}
}
\lstdefinestyle{mystyle}{
    backgroundcolor=\color{backcolour},   
    commentstyle=\color{codegreen},
    keywordstyle=\color{magenta},
    numberstyle=\tiny\color{codegray},
    stringstyle=\color{codepurple},
    basicstyle=\ttfamily\footnotesize,
    breakatwhitespace=false,         
    breaklines=true,                 
    captionpos=b,                    
    keepspaces=true,                 
    numbers=left,                    
    numbersep=5pt,                  
    showspaces=false,                
    showstringspaces=false,
    showtabs=false,                  
    tabsize=2
}

\newcommand{\hi}[1]{\vspace{.25em} \noindent {\bf #1} }

\newcommand{\oursys}{MinerU-Popo\xspace}
\newcommand{\ourbench}{PostDocBench\xspace}

\title{MinerU-Popo: Universal Post-Processing Model for Structured Document Parsing}

\author[1,*]{Bangrui Xu}
\author[2,*]{Ziyang Miao}
\author[1]{Xuanhe Zhou}
\author[3]{Yiming Lin}
\author[1]{Zirui Tang}
\author[2]{Xiaomeng Zhao}
\author[2]{Fan Wu}
\author[2]{Cheng Tan}
\author[1]{Fan Wu}
\author[2]{Bin Wang}
\author[2]{Conghui He}

\affiliation[1]{Shanghai Jiao Tong University}
\affiliation[2]{Shanghai Artificial Intelligence Laboratory, OpenDataLab}
\affiliation[3]{University of California, Berkeley}

\abstract{
VLM-based OCR models have become the de facto choice for document parsing, as they can accurately extract page-level elements (e.g., paragraphs within individual pages) together with their bounding boxes and textual content. However, downstream applications such as RAG require coherent document-level information, whereas these models often break cross-page continuity and fail to recover disrupted structures, such as paragraphs and tables truncated by page boundaries. Such relationships are not confined to a single page; instead, they require joint analysis of titles, paragraphs, tables, and images spanning multiple pages. A natural solution is therefore to reuse existing OCR outputs and reconstruct document-level logical structures through post-processing. 

To this end, we propose \oursys, a lightweight and universal framework for \textbf{PO}st-\textbf{P}rocessing \textbf{O}CR outputs, which converts page-level results from diverse parsers into coherent document-level structures. \oursys decomposes the problem into four focused subtasks: text truncation recovery, table truncation recovery, title hierarchy reconstruction, and image-text association. To address these effectively, we build a task-oriented data engine with task-specific input filtering, and use the generated data (\textbf{30K}) to fine-tune a \textbf{lightweight 4B post-processing model}. To support long documents, we introduce dynamic chunking with overlap-based synchronization, which aligns chunk-level outputs from the fine-tuned model and preserves global consistency. Finally, we assemble the aligned outputs into a tree-structured document representation, further enriched with node chunking and summaries for downstream retrieval and analysis. 
Empirical results show \oursys improves title-hierarchy TEDS by at least 20\% across all five tested OCR models, improves RAG accuracy on most subsets, and reduces per-query latency by up to 70\% compared with raw OCR baselines.
}

\contribution[*]{Equal contribution.}
\correspondence{Xuanhe Zhou, \email{zhouxuanhe@sjtu.edu.cn}, Bin Wang, \email{wangbin@pjlab.org.cn}, Conghui He, \email{heconghui@pjlab.org.cn}}

\metadata[Code]{\url{https://github.com/opendatalab/MinerU-Popo}}
\date{\today}

\begin{document}

\maketitle

\input{1_introduction}

\input{2_relatedwork}
\input{3_framework}
\input{4_experiments}
\input{5_conclusion}

\newpage

\bibliographystyle{IEEEtran}
\bibliography{reference}

\newpage
\appendix
\input{appendix}

\end{document}

%% file: 1_introduction.tex
\section{Introduction}
\label{sec:introduction}

\begin{figure}[!t]
    \centering
    \includegraphics[width=\linewidth]{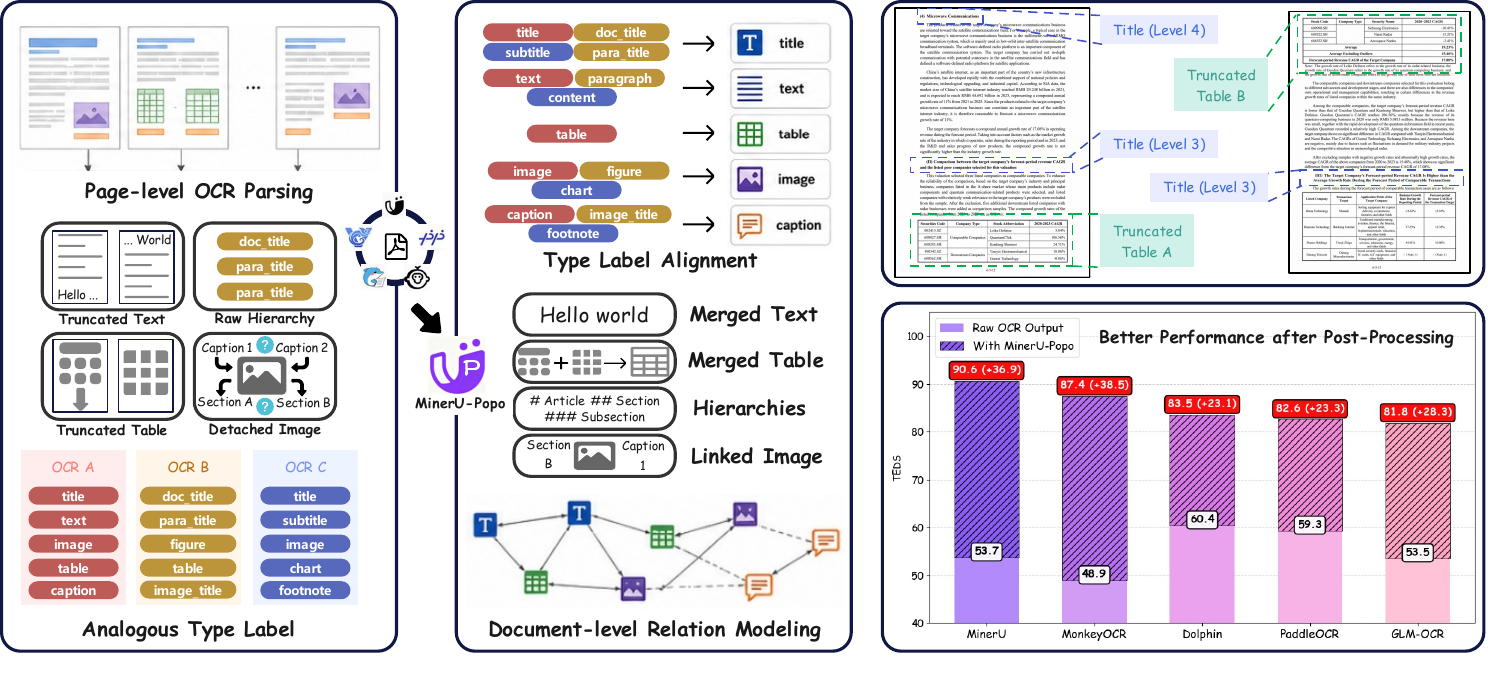}
    \vspace{-1.75em}
    \caption{\oursys supports diverse OCR models through type-label alignment and bridges page-level parsing with document-level modeling via four post-processing subtasks. Applying \oursys after OCR yields clear improvements in title hierarchy analysis.}
\vspace{-1.25em}
    \label{fig:intro}
\end{figure}

Document parsing converts unstructured and semi-structured documents into machine-readable representations that support downstream retrieval, question answering, and automated analysis~\cite{xu2026modora,zendb,sun2025docagent,saad2024pdftriage}. Recent VLM-based OCR models, such as MinerU~\cite{wang2026mineru}, PaddleOCR~\cite{cui2026paddleocr}, GLM-OCR~\cite{duan2026glm}, Dolphin~\cite{feng2026dolphin} and MonkeyOCR~\cite{zhang2025monkeyocrv15technicalreport}, have substantially improved page-level parsing by extracting element types, bounding boxes, reading order, and textual content from individual pages. However, downstream applications (e.g., RAG) rarely operate on isolated pages. They require coherent document-level structures, while page-wise OCR often breaks cross-page continuity and obscures relationships such as heading hierarchies, figure-caption links, and paragraphs or tables truncated by page boundaries. For example, Figure~\ref{fig:intro} shows a table split across two pages, where accurate reconstruction requires jointly analyzing cross-page table continuity and the surrounding title hierarchy.

% ~\cite{wang2026mineru,cui2026paddleocr,duan2026glm,dolphin2025,zhang2025monkeyocrv15technicalreport}

%Document parsing is a fundamental step for converting unstructured and semi-structured documents into structured data assets that can support downstream retrieval, question answering, and automated analysis~\cite{xu2026modora,zendb,sun2025docagent,saad2024pdftriage}). 
%Current document parsing methods primarily rely on specialized \zxh{vision language models (VLMs)} (e.g., MinerU~\cite{wang2026mineru}, PaddleOCR~\cite{cui2026paddleocr}, GLM-OCR~\cite{duan2026glm}, Dolphin~\cite{dolphin2025} and MonkeyOCR~\cite{zhang2025monkeyocrv15technicalreport}) to reconstruct page layouts. These models successfully achieve page-level parsing of element types, bounding box coordinates, and text content. 

A natural solution is therefore to reuse existing OCR outputs and reconstruct document-level logical structure through post-processing. This direction is appealing because modern OCR models already expose a broadly compatible intermediate representation: a reading-order sequence of typed blocks, each associated with content and coordinates. Moreover, these block types (e.g., text, title, image, table, caption) are largely aligned across systems and can often be normalized with lightweight rules. However, resorting to manual correction or large general-purpose VLMs for post-processing is costly for massive documents, difficult to scale, and often incompatible with privacy-sensitive or on-premise deployment~\cite{zhang2025docr, wang2026mineru}. For instance, according to telemetry from a public document parsing platform~\cite{wang2026mineru}, the service receives millions of OCR requests per month, with each document containing more than 20 pages on average.

However, realizing effective document post-processing is non-trivial. There are three fundamental challenges:
\textit{\textbf{(1) Cross-Page Geometric Discontinuity:}} Pagination breaks the document's logical flow, causing related elements to appear in separate pages, such as $(i)$ a section heading at the bottom of one page may govern content on the next page, and $(ii)$ a table may be split across two pages. 
\textit{\textbf{(2) Redundant Document Parsing:}} OCR parsers produce comprehensive but redundant outputs that include detailed information for all element blocks. In contrast, each post-processing subtask only depends on a small subset of relevant elements, while extraneous content may introduce interference. For instance, text truncation recovery should focus on paragraph fragments interrupted by page boundaries, rather than complete paragraphs or unrelated neighboring blocks. 
\textit{\textbf{(3) Scalability to Long Documents:}} Processing exceptionally long documents, such as those with hundreds of pages, in a single pass exceeds the context limits of lightweight models, while naive chunking can produce inconsistent hierarchy decisions across chunks.

To address these challenges, we introduce \oursys, a unified document post-processing framework that bridges OCR outputs and document-level structural parsing across diverse mainstream OCR models. Instead of treating OCR predictions as final parsing results, \oursys refines them through a locally deployed fine-tuned model that performs four document-structure subtasks: table truncation analysis, text truncation analysis, title hierarchy analysis, and image-text association analysis. To support robust structural reasoning, we construct training data that captures the diversity and representativeness of real-world complex document layouts, and further introduce task-specific data filtering to retain only the most critical evidence in the model input. To handle long documents, \oursys employs dynamic chunking and synchronization, which preserves global consistency while reducing deviations across independently processed chunks. Finally, \oursys enriches the refined parsing results through long-node chunking and summary generation, transforming fragmented OCR outputs into a unified, hierarchical, and semantically detailed document representation suitable for downstream retrieval and analysis. 

Our contributions are summarized as follows:

\noindent(1) We propose \oursys, a lightweight and universal framework for post-processing OCR outputs into coherent document-level structures.

\noindent(2) We develop a task-oriented data engine with task-specific input filtering, enabling a lightweight model to solve four focused cross-page structural parsing subtasks efficiently.

\noindent(3) We introduce dynamic chunking with overlap-based synchronization and an enriched tree-structured document representation, improving both long-document consistency and downstream retrieval quality.

\noindent(4) {Empirical results show \oursys generally improves OCR parsing result (e.g., more than 20\% hierarchy TEDS improvement for 5 OCRs), improved answering performance on RAG and End-to-End QA, and up to 70\% lower per-query latency of RAG.}

%% file: 2_relatedwork.tex
\section{Related Work}
\label{sec:related}

\hi{OCR Models.} 
%OCR旨在完成文本检测与识别，是文档解析的基础环节之一。传统阶段以手工特征与经典学习方法为主，在简单字符识别上取得了一定成效。深度学习兴起后，基于CNN或Transformer的识别器显著提升了多语言与复杂场景下的准确率。近年来，多模态大模型与任务专用模型的发展推动OCR从‘仅文字识别’到‘文档版面重建’的转变，相关模型（如MinerU, PaddleOCR, GLM-OCR）能力扩展到了元素类型分析、阅读顺序梳理、表格和公式格式识别等功能，在OmniDocBench等页面级别的文档解析数据集上均取得了很高的水平。
Recently, VLMs have driven a paradigm shift in OCR, extending beyond text recognition to comprehensive page layout reconstruction. Modern OCR models such as MinerU~\cite{wang2026mineru}, PaddleOCR~\cite{cui2026paddleocr}, GLM-OCR~\cite{duan2026glm}, Dolphin~\cite{feng2026dolphin}, and MonkeyOCR~\cite{zhang2025monkeyocrv15technicalreport} have established state-of-the-art page-level parsing performance on benchmarks like OmniDocBench~\cite{ouyang2025omnidocbench}.  % As a fundamental tool for document parsing, Optical Character Recognition (OCR) initially relied on handcrafted features~\cite{plotz2009markov}. Subsequently, CNNs~\cite{shi2016end} and Transformers~\cite{li2023trocr} significantly improved performance in complex settings. 
However, as compared in Table~\ref{tab:ocrsupport}, these models exhibit limitations in handling document-level relationships: \textit{\textbf{(1) Text Truncation:}} Most models detect intra-page truncations but fail to recover paragraphs broken by page boundaries. \textit{\textbf{(2) Title Hierarchy:}} GLM-OCR and Dolphin rely on rigid, predefined title levels (e.g., GLM-OCR distinguishes only two levels, \texttt{doc\_title} and \texttt{para\_title}, and Dolphin supports three), making them inadequate for documents with deeply nested hierarchical structures. In contrast, \oursys predicts open-ended levels to naturally accommodate arbitrary depths. \textit{\textbf{(3) Table Truncation:}} While PaddleOCR merges truncated tables, it cannot unify adjacent cells across page boundaries. \textit{\textbf{(4) Image-Text Association:}} None of the baselines capture semantic associations between images, captions, and sections. Ultimately, these deficiencies preclude the accurate parsing of multi-page documents. %, necessitating our proposed post-processing framework

\begin{table}[t]
\small
\centering
\caption{Current OCR Models' Capability to Handle Cross-page Relationships.}
\label{tab:ocrsupport}
\renewcommand{\arraystretch}{0.78}
\resizebox{\linewidth}{!}{%
\begin{tabular}{l|ccccc}
\toprule
\textbf{Model} & MinerU & Paddle & GLM-OCR & Dolphin & Monkey \\
\midrule
\textbf{Table Truncation} & Supported & Partial & Unsupported & Unsupported & Supported\\
\textbf{Text Truncation} & Partial & Partial & Partial & Unsupported & Partial\\
\textbf{Title Hierarchy} & Unsupported & Supported & Partial & Partial & Supported\\
\textbf{Image-Text Association} & Unsupported & Unsupported & Unsupported & Unsupported & Unsupported\\
\bottomrule
\end{tabular}%
}
\vspace{-1.5em}
\end{table}

\hi{Hierarchy Analysis.}
Hierarchy analysis models the nested structure of document elements. End-to-end approaches, such as DocHieNet~\cite{xing2024dochienet} and HRDoc~\cite{ma2023hrdoc}, infer relationships based on OCR outputs but underutilize explicit OCR priors (e.g., element types and reading order), forcing models to learn them implicitly. ZenDB~\cite{zendb} clusters elements using visual metadata (e.g., font size) from born-digital documents, which lacks generalizability to scanned documents and ignores semantic context. In contrast, \oursys explicitly exploits OCR-derived types and reading order to simplify model input via task-specific data filtering. By decomposing hierarchy analysis into focused subtasks (e.g., title structuring and image-text association), each operates exclusively on relevant elements, eliminating global processing overhead.

\hi{Document QA.}
%基于文档的问答是文档建模的重要下游任务。由于直接使用GPT, Gemini等VLM处理文档时存在性能随长度下降、调用开销大等问题，M3DocRAG,SV-RAG引入Colpali等基于视觉嵌入的搜索模型，检索文档中的相关页面后进行回答。而MoDora, ZenDB, DocAgent, PDFTriage等工作各自通过不同的解析和建模方法将文档转化为树或图结构,而后检索相关的节点用于分析。同时，现有的Vidore-v3, MP-DocVQA等数据集为文档问答提供了丰富的评测基准。
%Document-based question answering is an important downstream task in document modeling. Directly applying VLMs such as GPT and Gemini to process entire documents suffers from performance degradation and high inference cost with increasing input length and is not suitable for large scale documents processing. To address these issues, M3DocRAG~\cite{cho2024m3docrag} and SV-RAG~\cite{chen2024sv} incorporate image-embedding-based retrieval like Colpali~\cite{faysse2025colpali} to first retrieve relevant pages from the document before generating answers. Additionally, MoDora, ZenDB, DocAgent and PDFTriage ~\cite{xu2026modora, zendb, sun2025docagent, saad2024pdftriage} employ various parsing and modeling approaches to convert documents into tree or graph structures, followed by retrieving relevant nodes for analysis. In parallel, existing datasets like ViDoRe V3~\cite{loison2026vidore} provide comprehensive evaluation for document-based question answering.
Document-based question answering (QA) is a pivotal downstream task. Directly processing entire documents with Vision-Language Models is unscalable due to context-length degradation and prohibitive costs. Recent retrieval-augmented paradigms adopt two main trajectories: Vision-centric approaches (e.g., M3DocRAG~\cite{cho2024m3docrag}, SV-RAG~\cite{chen2024sv}) leverage image embeddings for page-level retrieval; Structure-aware frameworks~\cite{xu2026modora, zendb, sun2025docagent, saad2024pdftriage} model documents as trees or graphs for fine-grained, node-level retrieval. {Post-processing provides information necessary for document modeling, such as the image-text association and title hierarchy required by ZenDB~\cite{zendb} and MoDora~\cite{xu2026modora}, liberating them from the originally simplistic rule-based approaches.}
Meanwhile, benchmarks like ViDoRe V3~\cite{loison2026vidore} and MMDA~\cite{xu2026modora} have been introduced to rigorously evaluate these systems.

% Document-based question answering (QA) is a pivotal task in document understanding. Directly processing entire documents with Vision-Language Models (VLMs, e.g., GPT and Gemini) is unscalable due to context-length-induced performance degradation and prohibitive inference costs. To mitigate this, recent retrieval augmented paradigms adopt two main trajectories. Vision-centric approaches, such as M3DocRAG~\cite{cho2024m3docrag} and SV-RAG~\cite{chen2024sv}, leverage image embeddings (e.g., ColPali~\cite{faysse2025colpali}) for page-level retrieval prior to generation. Conversely, structure-aware frameworks~\cite{xu2026modora, zendb, sun2025docagent, saad2024pdftriage} model documents as trees or graphs to enable fine-grained, node-level retrieval. Concurrently, comprehensive benchmarks like ViDoRe V3~\cite{loison2026vidore} and MMDA~\cite{xu2026modora} have been introduced to rigorously evaluate these evolving systems.

%% file: 3_framework.tex
\section{Problem Formulation}
\label{subsec:problem}

\begin{figure}[t]
    \centering
    \includegraphics[width=.95\linewidth]{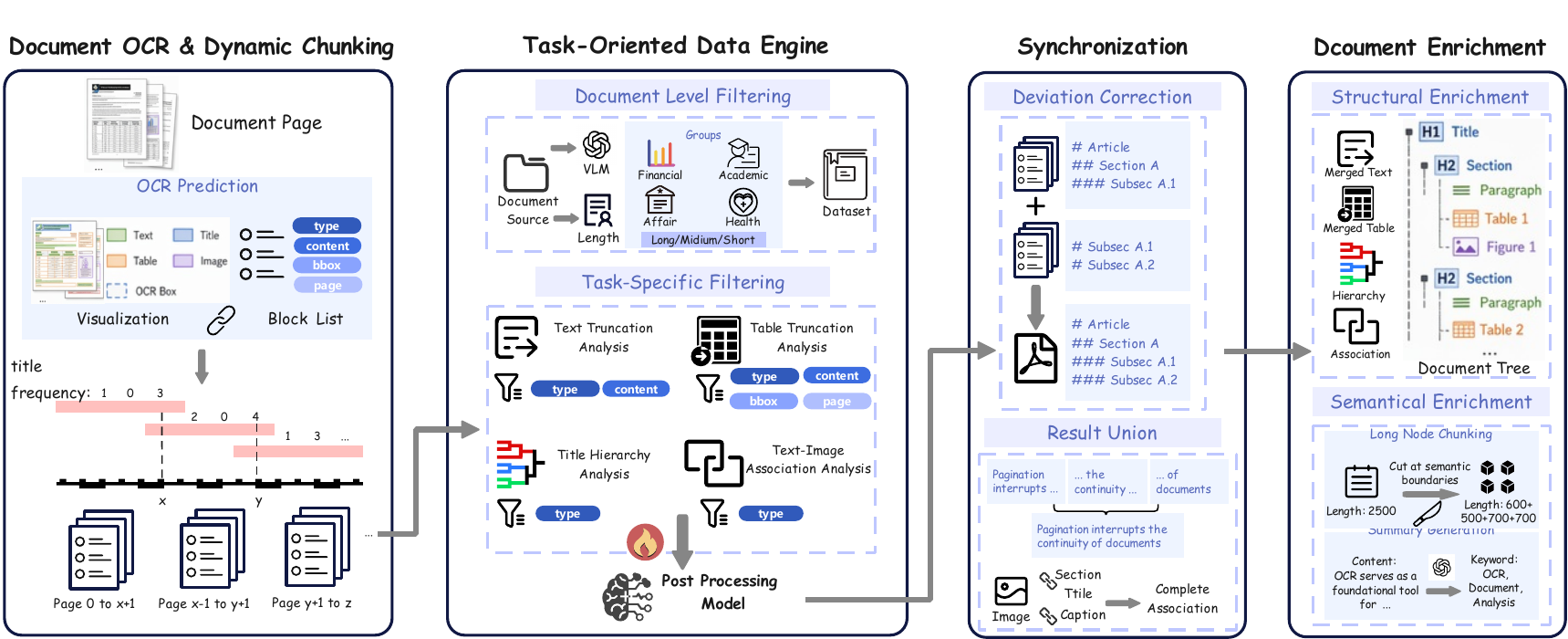}
    %\vspace{-1.25em}
    \caption{Overview of \oursys. The post-processing model training and inference are supported by a task-oriented data engine that collects complex and representative documents as datasets and simplifies the model input for each subtask. Page-level OCR predictions of long documents are grouped via dynamic chunking, and then synchronized after chunk-level post-processing. The result is enriched via structural tree construction as well as semantic chunking and summarization.}
    \label{fig:overview}
    \vspace{-.75em}
\end{figure}

% \hi{Overall Goal.}
The goal of post-processing is to bridge this gap by taking a sequence of OCR-extracted block elements $\mathcal{E} = \{e_1, e_2, \dots, e_n\}$ as input, and outputting a coherent, document-level structured tree $\mathcal{T}$. Crucially, this process must address the challenges of cross-page discontinuity and logical structure absence, while strictly balancing \textit{\textbf{Accuracy}} (precisely recovering truncated elements and deep hierarchies) and \textit{\textbf{Efficiency}} (minimizing computational overhead by filtering redundant OCR information). To achieve this, we decompose the overall post-processing into four focused subtasks.

\hi{Table Truncation Analysis.}
Pagination often splits a single logical table into multiple fragments across pages. For any two adjacent table elements $e^{table}_i$ and $e^{table}_{i+1}$ across a page boundary, this subtask first needs to predict a table-level continuation label $p_i \in \{0,1\}$ to indicate whether they belong to the same logical table. If $p_i=1$, we need to further predict a column-wise cell merging strategy $Q_i=\{q_1,\ldots,q_x\}$, where $q_j \in \{0,1\}$ indicates whether the boundary cells in column $j$ should remain separate (simple concatenation) or be merged into one continuous cell.

\hi{Text Truncation Analysis.}
Similarly, paragraphs are frequently interrupted by page breaks, column breaks, or inserted figures. For a sequence of text elements, this subtask needs to predict a binary label $p_i \in \{0,1\}$ for each adjacent text pair $(e^{text}_i, e^{text}_{i+1})$ to determine whether they are logically truncated and should be concatenated into a single continuous paragraph.

\hi{Title Hierarchy Analysis.}
To reconstruct the document's structural backbone, we must determine the nesting levels of all section headings. Given a sequence of title elements extracted from the document, this subtask needs to predict an open-ended integer level for each title (i.e., $e^{title}_i.level \in \{1, 2, 3, \dots\}$), where a smaller number represents a higher hierarchical level (e.g., Level 1 for document title, Level 2 for main sections).

\hi{Image-Text Association Analysis.}
Visual elements must be grounded in their semantic context. For each caption element $e^{cap}_i$, this subtask needs to predict its associated image or table. Furthermore, to embed these visual elements into the document tree, it predicts the corresponding section title $e^{title}_j$ that logically governs each image or table.

\section{Methodology} 
\label{sec:framework}

As illustrated in Figure~\ref{fig:overview}, \oursys presents a comprehensive pipeline to transform fragmented OCR outputs into an accurate document-level representation. The pipeline begins with a lightweight model fine-tuned on a \textbf{Task-Oriented Data Engine}, which filters out redundant OCR noise to provide focused inputs for each of the four subtasks mentioned above. To efficiently handle arbitrarily long documents without losing global context, we employ a \textbf{Dynamic Chunking and Synchronization} strategy during inference. Finally, through \textbf{Document Enrichment}, the refined elements are assembled into a hierarchical tree structure, further optimized with semantic chunking and LLM-generated summaries to facilitate downstream retrieval and analysis.

 % \oursys employs a model fine-tuned on four post-processing tasks over documents with diverse and complex layouts, where each subtask filters the most relevant information from OCR predictions as its input. If the document is extremely long, \oursys dynamically splits it into several page chunks, processes them one by one and synchronizes the results. Finally, \oursys performs the document enrichment based on model outputs. Structurally, it defines nodes and edges to construct a tree structure. Semantically, it divides long paragraph nodes at semantic boundaries, and generate summaries for each node.

 % \oursys utilizes a model fine-tuned on four post-processing tasks to handle complex document layouts, where each subtask filters specific elements from OCR predictions as input. For lengthy documents, it dynamically splits pages into chunks for processing and synchronization. Finally, it enriches the document representation both structurally, by constructing a document tree, and semantically, by segmenting long sections at semantic boundaries and generating node-level summaries.

\subsection{Task-Oriented Data Engine}
\label{subsec:taskdata}

%Raw OCR outputs contain diverse multi-modal elements (text, images, tables) spanning the entire document. Simply concatenating them as model input introduces extraneous noise, dilutes critical information, and exacerbates long-context reasoning challenges. \textbf{Technical Strategy:} 
We first provide a task-oriented data engine that applies task-specific input filtering. By retaining only the crucial element types and structural priors relevant to each specific subtask, we significantly reduce input complexity and enhance the model's focus.

\hi{Document-Level Filtering.}
We collect representative and diverse documents from large-scale real-world documents including Common Crawl (CC), a neutral, non-profit public web archive. Following the processing recognized by the academic community~\cite{laurencon2023obelics}, we ensure that the CC data is used solely for non-commercial academic research. We strictly adhere to the terms of use~\cite{ccuse} and guarantee compliance with the Robots protocol. For all collected documents, we apply a pre-filtering step to remove those containing private or sensitive information.

To obtain a balanced document distribution, we first use a VLM to categorize these documents according to their contents (e.g., technology, academic, education, healthcare, affair, and law), layout characteristics, visual styles, and related attributes. We then group the documents by length and sample from each category to achieve balanced coverage across both document type and length. 

Document OCR yields multi-modal elements (text, images, tables), and simply concatenating them as input introduces extraneous noise and exacerbates long-context challenges (e.g., tables are irrelevant for determining title hierarchies). By decomposing the overall task (Section~\ref{subsec:problem}), each subtask only requires specific element types. Thus, we employ task-specific filtering during training and inference to retain crucial relevant elements and pages, significantly reducing input complexity.

\hi{Task-Specific Filtering.} For each subtask, we filter the most relevant elements as input via heuristics.

\textit{(1) Filtering for Hierarchy Analysis.} Titles play a governing role over text paragraphs. Once the hierarchical levels of titles are determined, the hierarchy of the text subordinate to each title is consequently established. Therefore, in title hierarchy analysis, only elements of type ``title'' are filtered out and arranged according to the reading order predicted by OCR as the input.
%Yiming的comment是否认为这里我们是逐个处理每个标题？补充了第三章的定义以明确这里的任务是批次处理，因此是有上下文标题信息的。

\textit{(2) Filtering for Image-Text Association Analysis.}
Images and tables are, on one hand, associated with their respective captions, and on the other hand, subordinate to specific sections (represented by their titles) within the document. To capture these associations, we filter elements of type title, caption, image or table, discarding the massive volume of regular body text.

%To capture the associations between visual elements and their textual context, the model only requires elements that participate in these relationships. we selectively filter and retain titles, captions, images, and tables, discarding the massive volume of regular body text.

\textit{(3) Filtering for Text Truncation Analysis.}
%Text elements actually constitute a large portion of the document, and even after filtering by type, their content remains excessively lengthy. Since truncation occurs only at the beginning or end of each OCR-output text block, we further exploit this property to filter and simplify each text element. Specifically, we first apply heuristic rules based on the starting patterns of the text (e.g., prefixes such as 'a.', '1.2') and terminating symbols (e.g., periods or question marks at the end) to exclude text blocks that exhibit no truncation at their beginning or end. Then, for the remaining text element boundaries where truncation may exist, we extract only the first and last sentences of long paragraphs based on the terminating symbols, which serve as the final input text.
Text elements dominate document length. However, truncation inherently occurs only at the boundaries of text blocks. We leverage this structural prior by applying heuristic rules based on starting prefixes (e.g., '1.1') and terminating punctuation (e.g., periods) to discard explicitly untruncated blocks. For the remaining candidates, we extract only the first and last sentences as the final input, drastically reducing token count.

% Text elements dominate document content and remain prohibitively long after type-based filtering. Since OCR truncation occurs at the boundaries of text blocks, we leverage this structural prior for further simplification. Specifically, we apply rules based on starting prefixes (e.g., 'a.', '1.1') and terminating punctuation to discard explicitly untruncated text blocks. For the remaining text blocks, we extract only the first and last sentences as the final input.

\textit{(4) Filtering for Table Truncation Analysis.}
%Table truncation analysis针对每个合并表格，细粒度分析每列的合并方式。而布局中的表格截断的因素主要来自跨页，因此首先利用版面位置、续表标记、表格宽度以及结构一致性规则来筛选出那些需要合并的表格。具体来说，我们要求两个候选表格必须出现在相邻页面的交界处，即按阅读顺序，当前页的第一个元素和上一页的最后一个元素类型均为表格。其次，检查表格的caption信息，根据是否有各自的caption以及"continue"等续表标识，确定二者是否为一个整体表的截断结果。进一步，从视觉角度比较两个表格的bbox宽度，计算其差值是否在一个允许的阈值中。最后，比较表头，列数等结构信息，要求二者存在相同的表头或者列数。
%For each merged table, we perform a fine-grained analysis of the merging pattern of each column. Since table truncation in the layout primarily arises from page breaks, we first exploit layout position, continuation table markers, table width, and structure consistency heuristics to predict truncated tables. Concretely, we require that two candidate tables appear at the boundary of adjacent pages. That is, in reading order, the last element of the current page and the first element of the next page must both be a table. Next, we examine the caption information of the tables based on the presence of their respective captions and continuation indicators such as "continued". Furthermore, from a visual perspective, we compare the bounding box widths of the two tables and compute whether their difference falls within a permissible threshold. Finally, we check structural information, requiring alignment of two tables on their column names and numbers.
Table truncation typically results from page breaks. To identify candidates efficiently without heavy model inference, we employ layout and structural heuristics. We select tables located at page boundaries (the last element of one page and the first of the next). We then validate these candidates by checking for continuation markers (e.g., ``continued'' in captions), bounding box width consistency, and identical column counts.

% Since table truncation typically results from page breaks, we identify truncated candidates using layout, textual, visual, and structural heuristics. Specifically, candidate tables must be at page boundaries, acting as the last and first elements of consecutive pages. We then validate these candidates by checking for continuation markers (e.g., "continued") in their captions, the bounding box widths (should align within a predefined threshold), as well as structural consistency (identical column counts and column names).

%\hi{Filtering for Visual Input.} 
%In addition to the above mentioned element information described by textual modality, input in visual modality also provides important information for each subtask. For instance, the size and font of titles can reflect certain hierarchical relationships, the page image can directly reveal factors that lead to truncation of text paragraphs, and the contents of images are also concretely presented by visual modality. To fully leverage the aforementioned information presented by images, for each subtask except table truncation analysis, we extract document pages that contain the filtered elements, label the corresponding page number on the image of each page, and concatenate all page images as the visual modality input to the model.

\subsection{Dynamic Chunking and Synchronization}
\label{subsec:chunk}

relevant elements, processing long documents remains a bottleneck. For instance, a Bank of America financial report spans hundreds of pages and thousands of paragraphs, making single-pass processing prohibitive. This necessitates segmenting the document into manageable page chunks.
However, naive chunking introduces inconsistencies in hierarchical predictions. For instance, text truncation at chunking boundaries can not be detected. Another case is that the global document title (level 1) is localized to the first chunk, and subsequent chunks lacking this context can erroneously promote section titles from level 2 to 1. To mitigate these artifacts, we propose a dynamic chunking and synchronization strategy.

\hi{Dynamic Chunking.} %To prevent the loss of truncations at boundaries and provide reference information for synchronization, we first adopt an overlapping chunking strategy, where adjacent chunks are required to share a three-page overlap. Furthermore, to provide richer reference information for subsequent synchronization, chunk boundaries are not computed using a fixed stride. Instead, based on a preset stride and threshold, we dynamically select as boundaries the pages with the highest frequency of task-specific element types (e.g., 'title' type for title hierarchy analysis) within a certain range. This mechanism first ensures the presence of required elements within the overlapping region, which is necessary for calculating the deviation in the subsequent synchronization step. And maximizing the occurrence frequency of required elements is to reduce the randomness in deviation calculation due to few reference elements. The overall process can be summarized as following formulas:
To prevent information loss from boundary truncation and establish reference points for synchronization, we employ an overlapping chunking strategy with a three-page overlap. Rather than using fixed strides, we dynamically determine chunk boundaries to enrich reference information. Specifically, within a search window defined by a preset stride and threshold, we select the page with the highest frequency of task-specific elements (e.g., 'title' for hierarchy analysis) as the boundary. This density-maximizing mechanism guarantees sufficient reference elements in the overlapping regions, thereby reducing variance in the subsequent deviation calculation during synchronization. Formally, this process is defined as:
\[ b_0 = 0, b_i = p|f(p,e) >= f(p_0,e) \enspace for \enspace \forall p_0 \in [b_{i-1} + s - t, b_{i-1} + s + t] \]
\[ chunk_i = (max(0,b_i -1), min(b_{i+1}+1, p_{max})) \]

\hi{Chunk Synchronization.} %The synchronization stage collects the model output of each chunk, and integrate them to produce a unified final output. For predicted title levels for each chunk by title hierarchy analysis, the deviation caused by discrete title distribution is reflected by different prediction for title levels in the overlap of two chunks. Using the first chunk as the reference, we recursively adjust the level of each subsequent chunk from front to back, based on its level deviation relative to the preceding chunk within the overlapping region. The overall process can be summarized as following formulas:
The synchronization stage aggregates chunk-level outputs into a unified document-level prediction. Due to discrete title distributions, hierarchical predictions often diverge within overlapping regions. Using the initial chunk as an anchor, we sequentially calibrate the hierarchy levels of subsequent chunks based on their average deviations compared to the former chunk at their overlap. This process is formalized as follows:
\[ level_{i,x} = pred(chunk_i,title_x), level_{i+1,y} = pred(chunk_i,title_y) \]
\[ deviation = avg(level_{i,a}-level_{i+1,a}), title_a \in overlap\]
\[ level_{i+1,y}^{final} = level_{i+1,y} + deviation \]

For other subtasks that do not involve such deviations, the synchronization is relatively simple, and the final output is the union of each chunk's result.

\subsection{Document Enrichment}
\label{subsec:tree}

% The raw post-processing outputs are flat relational labels. For downstream tasks like RAG, documents must be organized into a searchable structure. Furthermore, embedding full-text nodes incurs high computational overhead, while embedding only short titles lacks sufficient semantic granularity.
% \textbf{Technical Strategy:} 

We perform structural enrichment to enrich both the structural and semantical document-level information via hybrid node chunking and LLM-generated summarization.

\hi{Structural Enrichment.}
Based on the predicted title hierarchies and associations, we assemble elements into a tree $\mathcal{T}$. The hierarchical levels determine the parent-child relationships of section nodes, while text paragraphs, merged tables, and associated images are attached as child elements to their respective governing section nodes.

\hi{Semantical Enrichment.}
%The primary tree derived from the above mentioned algorithm has effectively captured the document structure, but is not yet suitable for downstream retrieval and analysis tasks. First, tree nodes are partitioned by titles, which implies that for documents with sparse titles (e.g., novels and proses), the text within each node remains excessively long, exceeding the length for effective comprehension and analysis. Second, retrieval based on vector matching using the full content of nodes incurs high computational overhead of embeddings, whereas the node title is too coarse-grained to serve as a basis for retrieval. For instance, a title "Methodology" alone provides no indication of framework design. Thus, we conduct node chunking and summary generation to solve these problems.
While the derived primary tree captures document structure, it remains suboptimal for downstream retrieval and analysis. First, title-based partitioning yields excessively long nodes for documents with sparse titles (e.g., novels or prose), hindering effective processing. Second, full-content embedding incurs much overhead, yet isolated titles (e.g., "Methodology") lack sufficient granularity for accurate retrieval. To address these limitations, we introduce node chunking and summary generation. During node chunking, we sequentially process each node's text. Once the accumulated text length reaches the threshold, we segment it into a sub-node at the next paragraph boundary. These boundaries are derived from OCR outputs, but truncation boundaries $({e^{text}_i}, {e^{text}_{i+1}})$, $p_{i} = 1$ are excluded. 
Following node chunking, we prompt an LLM to generate summaries for all nodes. The resulting document tree balances optimal granularity with information completeness. Serving as an intermediate representation between titles and full texts, these summaries distill crucial arguments, methods, trends, or conclusions while abstracting away details.

% \xbr{This strategy fully integrates the length thresholds, OCR paragraph segmentation, and text truncation analysis, achieving more reasonable chunking.}

%% file: 4_experiments.tex
\section{Experiments}
\label{sec:experiments}

\subsection{Test Datasets} % for Document Post-Processing

We construct a dataset with 30K training instances and 165 test instances for each subtask to evaluate post-processing performance (\ourbench). Existing OCR benchmarks, such as OmniDocBench~\cite{ouyang2025omnidocbench}, DocLayNet~\cite{pfitzmann2022doclaynet}, olmOCR-Bench~\cite{poznanski2025olmocr}, are not suitable for this evaluation because they mainly target page-level parsing, whereas our focus is document-level post-processing. For downstream tasks, we adopt ViDoRe V3~\cite{loison2026vidore} for Retrieval-Augmented Generation (RAG) evaluation, utilizing its bounding box annotations for spatial retrieval accuracy. Additionally, we use MMDA~\cite{xu2026modora} for end-to-end QA evaluation to assess the representational capacity of different document structures. In all experiments, we utilize MinerU~\cite{wang2026mineru} as the base OCR model, fine-tune Qwen3-VL-4B~\cite{qwen3technicalreport} for post-processing, and leverage Gemini-3-Flash~\cite{pichai2025new} for answer generation. % , including bounding boxes, element types, text extraction, and table or formula reconstruction

\subsection{Experiment Setup}

First, we support multiple mainstream OCR models, including MinerU~\cite{wang2026mineru}, PaddleOCR~\cite{cui2026paddleocr}, GLM-OCR~\cite{duan2026glm}, Dolphin~\cite{feng2026dolphin}, and MonkeyOCR~\cite{zhang2025monkeyocrv15technicalreport}. Besides, we fine-tune Qwen3-VL-4B~\cite{qwen3technicalreport} for post-processing, utilize Qwen3-Embedding-4B~\cite{qwen3embedding} for embedding-based retrieval, and leverage Gemini-3-Flash~\cite{pichai2025new} for final answer generation. The training is conducted on 8 H200 GPUs for 6 hours.

\subsection{Experimental Results} 
\label{subsec:mainres}

\begin{table}[!t]
\centering
\caption{Model Performances Comparison on \ourbench.}
\label{tab:modelcompare}
\begin{tabular}{l|ccccc}
\toprule
\textbf{Model} & \oursys & Qwen3-VL-2B & Qwen3-VL-4B & Qwen3-VL-8B & Qwen3-VL-32B \\
\midrule
\textbf{TEDS}$\uparrow$ & \textbf{90.6} & 21.2 & 56.5 & 65.9 & 78.0\\
\textbf{Doc/s}$\uparrow$ & \textbf{0.37} & 0.22 & 0.20 & 0.16 & 0.04\\
\bottomrule
\end{tabular}%
\end{table}

\begin{figure}[t]
    \centering
    \includegraphics[width=\linewidth]{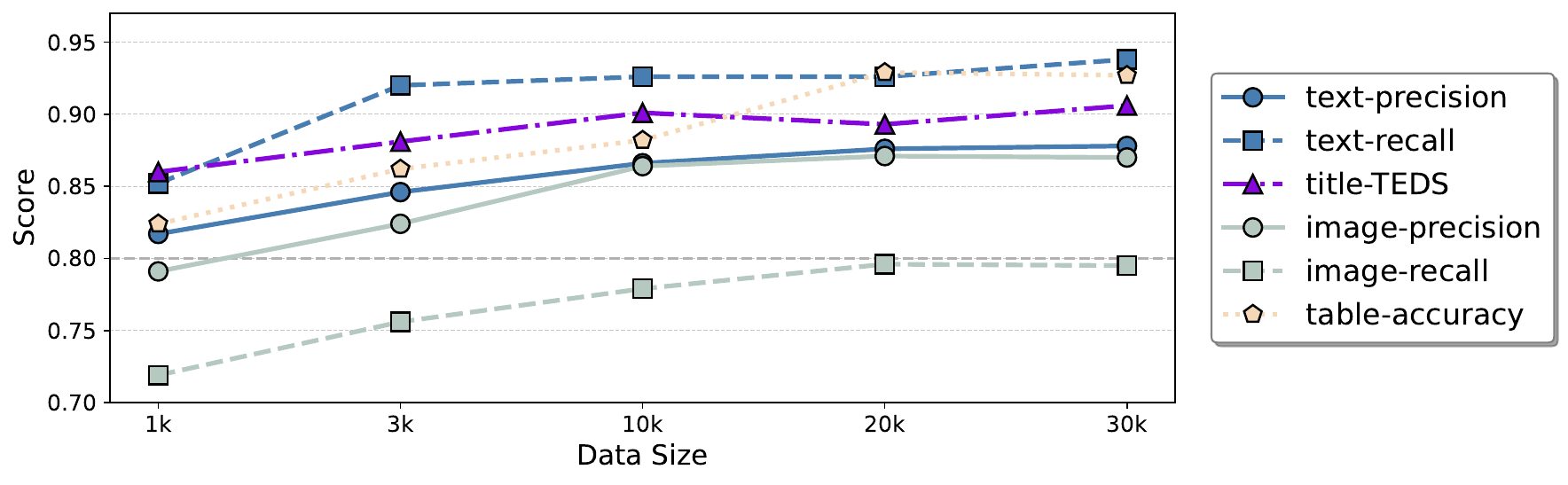}
    \caption{Performance on \ourbench with Different Training Data Size.}
    \label{fig:train}
\end{figure}

\textbf{Finding 1: \oursys achieves superior accuracy and inference speed compared to baselines, ultimately scoring around 90\% across diverse post-processing subtasks in fine-tuning supported by the task-oriented data engine.}

\hi{Post Processing Performance.} Table~\ref{tab:modelcompare} compares the TEDS score and inference speed, measured in documents per second, of \oursys and other pre-trained models on title hierarchy analysis under the same deployment conditions. \oursys achieves both a higher TEDS score than larger models, 90.6 vs. 78.0, and higher speed than even smaller models, 0.37 vs. 0.22. This shows that our task-oriented data engine and dynamic chunking improve both model training and inference efficiency by reducing input and output length.

We further evaluate the efficacy of fine-tuning by jointly training on varying subsets (1k to 30k instances per subtask) from \ourbench. As Figure~\ref{fig:train} illustrates, when training instances for each subtask increases from 1k to 10k, all the corresponding scores  improve. But as the data size expands from 10k to 30k, the performance gain diminishes substantially. This indicates that the current training data are sufficient, and more training data would yield only marginal benefits. 

For \oursys fine-tuned on the complete training set, the precision and recall for text truncation analysis reach 87.8\% and 93.8\%, those for image-text associations analysis are 87.0\% and 79.5\% respectively, while the scores of other two subtasks exceed 90\% (e.g., the TEDS between predicted and annotated title hierarchy is 90.6\%, the accuracy for merge method prediction in table truncation analysis is 92.7\%). This demonstrates that the fine-tuned model can make accurate judgments across various post-processing subtasks, allowing us to integrate its inference results to obtain a well-formed document modeling structure.

\textbf{Finding 2: \oursys improves retrieval and answering accuracy while reducing latency, highlighting the utility of post-processing in downstream tasks. The raw and visual RAG scored higher in some rare cases due to queries targeting very detailed content or complex tables.}

\hi{Comparison on RAG.} To evaluate the downstream utility of structured document modeling, we compared Retrieval-Augmented Generation (RAG) using \oursys to process document trees against raw OCR and visual retrieval baselines. Experiments are conducted on the five English subsets of ViDoRe V3. For tree-based retrieval, embeddings are computed solely on the concatenated title path and summary of each node. Lacking structure, the raw OCR baseline embeds full text, while the visual baseline retrieves relevant pages using ColEmbed-3B-v2~\cite{moreira2026_nemotron_colembed_v2}. Finally, the retrieved contexts (nodes, text, or page images) are fed to Gemini-3-Flash for question answering.

\begin{table}[t]
\centering
\caption{RAG Performance on ViDoRe V3.}
\label{tab:rag}
\resizebox{\linewidth}{!}{%
\begin{tabular}{l|ccc|ccc|ccc|ccc|ccc}
\toprule
\textbf{Subset} & \multicolumn{3}{c}{\textbf{C.S.}} & \multicolumn{3}{c}{\textbf{Fin.}} & \multicolumn{3}{c}{\textbf{H.R.}} & \multicolumn{3}{c}{\textbf{Ind.}} & \multicolumn{3}{c}{\textbf{Phar.}} \\
\midrule
\textbf{Metric} & Acc $\uparrow$ & Recall $\uparrow$ & Time $\downarrow$ & Acc $\uparrow$ & Recall $\uparrow$ & Time $\downarrow$ & Acc $\uparrow$ & Recall $\uparrow$ & Time $\downarrow$ & Acc $\uparrow$ & Recall $\uparrow$ & Time $\downarrow$ & Acc $\uparrow$ & Recall $\uparrow$ & Time $\downarrow$ \\
\midrule
\textbf{\oursys} & \textbf{84.4} & 45.7 & 84 & 49.5 & 40.5 & 68 & \textbf{66.8} & 47.5 & 15 & 58.7 & 35.6 & 26 & \textbf{71.6} & 44.3 & 9 \\
\textbf{raw RAG} & 82.3 & 39.7 & 145 & 48.7 & 32.4 & 225 & 63.2 & 41.6 & 39 & \textbf{60.4} & 37.1 & 53 & 64.4 & 28.6 & 20 \\
\textbf{visual RAG} & 80.7 & 80.3 & 129 & \textbf{58.4} & 41.1 & 273 & 64.8 & 69.4 & 104 & 59.7 & 58.4 & 481 & 67.6 & 69.6 & 214 \\
\bottomrule
\end{tabular}%
}
\end{table}

\begin{table}[t]
\centering
\caption{Distribution of Question Types}
\label{tab:qatypes}
\resizebox{\linewidth}{!}{%
\begin{tabular}{lccc|ccc|ccc|ccc|ccc}
\toprule
\textbf{Subset} & \multicolumn{3}{c}{\textbf{C.S.}} & \multicolumn{3}{c}{\textbf{Fin.}} & \multicolumn{3}{c}{\textbf{H.R.}} & \multicolumn{3}{c}{\textbf{Ind.}} & \multicolumn{3}{c}{\textbf{Phar.}} \\
\midrule
\textbf{Category} & 1 & 2 & 3 & 1 & 2 & 3 & 1 & 2 & 3 & 1 & 2 & 3 & 1 & 2 & 3 \\
\midrule
\textbf{Proportion(\%)} & 0.5 & 74.4 & 25.1 & 2.3 & 41.1 & 56.6 & 2.8 & 61.3 & 35.8 & 1.4 & 35.0 & 63.6 & 2.2 & 76.9 & 20.9 \\
\bottomrule
\end{tabular}%
}
\end{table}

Table~\ref{tab:rag} reports the accuracy, bounding box (bbox) recall, and latency (seconds/query) for RAG across subsets. Accuracy is evaluated via an LLM (scored 1 for completely, 0.5 for partially, and 0 for incorrectly answered), while bbox recall is the percentage of the overlapping area of the retrieved and annotated evidence relative to the annotated evidence area. \oursys outperforms RAG on raw OCR in four of the five subsets (e.g., improving accuracy by 7.15\% and bbox recall by 15.72\% on the Phar. subset). Furthermore, \oursys significantly reduces latency, cutting time costs by roughly 70\% and 50\% on the Fin. and H.R. subsets, respectively, by efficiently embedding concise node titles and summaries.
But relying on summaries for retrieval can miss fine-grained evidence, explaining its underperformance on the Ind. subset. 

To further illustrate the distinctions among title, summary, and content, we employ an LLM to classify ViDoRe V3 questions. The first category queries by exact section titles or captions of images, tables, and charts. The second category includes queries targeting crucial arguments, methods, trends, and conclusions. The third category focuses on specific numerical values and highly detailed content. Table~\ref{tab:qatypes} shows that among questions directed at real-world documents, the first category constitute a small fraction (less than 3\%). The second category (targeting crucial arguments, methods, trends, and conclusions) dominates, comprising >74.4\% of the C.S. and Phar. subset and >35.0\% of Ind. subset. By distilling crucial information, summaries enable efficient retrieval without repeatedly processing lengthy texts. The third category accounts for 20.9\%-63.6\% across subsets, where summaries yield marginal retrieval benefits since they capture high-level abstractions rather than fine-grained details.

Results in Table~\ref{tab:rag} align with the category distributions in Table~\ref{tab:qatypes}. Specifically, 63.6\% of Ind. queries require precise numerical values or fine-grained details that are often absent from high-level summaries, leading to lower accuracy when retrieval relies on node paths and summaries. In contrast, this mechanism remains effective for the other four subsets. These observations suggest that further gains may be achieved by selecting RAG strategies based on LLM-predicted query types.

Notably, while visual RAG achieves higher recall, \oursys yields superior overall answer accuracy. Visual RAG's high recall stems from page-level retrieval, which bypasses bounding box alignment and guarantees 100\% content recall upon a successful page match. However, deriving answers directly from high-resolution, text-dense page images strains the model’s visual comprehension, leading to lower accuracy in most subsets. An exception is the Fin. subset, where visual RAG achieves better answering accuracy. This is because over 25\% of its queries rely on complex financial tables (versus 10\% and simpler tables elsewhere). OCR struggles to parse such complex tables into OSTL or HTML format, and LLMs subsequently fail to reason over them. In contrast, VLMs process complex tables more effectively in their native image format. In addition to the retrieval recall and answering accuracy, visual RAG incurs substantial computational overhead in its page image embedding.

\textbf{Finding 3: JSON and Markdown formats derived from \oursys better capture document information for comprehensive question answering.}

\hi{Comparison on End-to-End Answering.}
To evaluate the representational capacity of different data formats, we conduct end-to-end question answering on MMDA across various evidence formats (Table~\ref{tab:mmdares}). JSON and Markdown are derived from our constructed document tree, whereas SHT and XML follow ZenDB~\cite{zendb} and DocAgent~\cite{sun2025docagent}, respectively.

\begin{table}[t]
\centering
\caption{End-to-End Answering Performance on MMDA.}
\label{tab:mmdares}
\renewcommand{\arraystretch}{0.8}
\begin{tabular}{l|ccccc}
\toprule
\textbf{Type} & \textbf{Overall}$\uparrow$ & \textbf{Text}$\uparrow$ & \textbf{Hybrid}$\uparrow$ & \textbf{Hierarchy}$\uparrow$ & \textbf{Location}$\uparrow$\\
\midrule
\textbf{JSON} & 64.2 & 70.7 & 58.6 & 71.1 & \textbf{61.6}\\
\textbf{Markdown} & \textbf{65.5} & \textbf{75.5} & \textbf{63.2} & \textbf{81.5} & 41.7\\
\textbf{SHT} & 60.2 & 71.0 & 54.6 & 62.0 & 49.6\\
\textbf{XML} & 52.2 & 64.1 & 42.8 & 58.5 & 40.1\\
\bottomrule
\end{tabular}%
\end{table}

We observe that tree-derived JSON and Markdown formats yield superior representations for document QA. JSON provides the most comprehensive representation by encoding both the semantics and coordinate location of document elements, achieving over 60\% accuracy on MMDA. However, locations become redundant for purely semantic queries and can even impede model comprehension. Consequently, JSON underperforms Markdown on location-agnostic questions, as Markdown is more amenable to semantic reasoning. Conversely, this omission also causes Markdown to suffer a 20\% performance drop on queries based on element locations. Finally, although SHT and XML are also tree-structured, they lack essential post-processing to resolve content truncations, hierarchies, and associations. This yields inferior overall performance, scoring only 50\% and 60\% on cross-modal and hierarchical questions respectively.

\subsection{Ablation Study}

\begin{table}[!t]
\centering
\caption{Ablation Study of Dynamic Chunking and Synchronization.}
\label{tab:ablation1}
\renewcommand{\arraystretch}{0.8}
\begin{tabular}{l|ccccc}
\toprule
\textbf{Metric} & text-pre$\uparrow$ & text-Rec$\uparrow$ & title-TEDS$\uparrow$ & image-pre$\uparrow$ & image-rec$\uparrow$ \\
\midrule
\textbf{w/} & 87.8 & 93.8 & 90.6 & 87.0 & 79.5\\
\textbf{w/o} & 84.0 & 86.6 & 85.8 & 83.7 & 75.2\\
\bottomrule
\end{tabular}%
\end{table}

\hi{Dynamic Chunking and Synchronization.}
Table~\ref{tab:ablation1} evaluates post-processing performance with and without dynamic chunking and synchronization. Processing long documents entirely degrades performance obviously (e.g., text truncation recall drops from 93.8\% to 86.6\%). This highlights the model's limited long-context reasoning capacity, motivating our chunk-based approach.

\hi{Semantic Enrichment.}
Table~\ref{tab:ablation2} ablates node chunking and summary generation in semantic enrichment. Here, IoU denotes the intersection over union between the retrieved and gold evidence bounding boxes. Replacing node summaries with full detailed contents yields higher overall IoU and Recall, but greatly increases the time cost. For instance, per-query latency on the Fin. subset jumps from 68s to 248s. Conversely, omitting node chunking accelerates retrieval due to a reduced node count. However, while larger unchunked nodes preserve more context, they introduce irrelevant noise that degrades the precision. Consequently, this omission decreases IoU across all subsets and lowers Recall on the C.S. and H.R. subsets.

\begin{table}[!t]
\centering
\caption{Ablation Study of Semantic Enrichment on ViDoRe V3 Retrieval.}
\label{tab:ablation2}
\resizebox{\linewidth}{!}{%
\begin{tabular}{l|ccc|ccc|ccc|ccc|ccc}
\toprule
\textbf{Subset} & \multicolumn{3}{c}{\textbf{C.S.}} & \multicolumn{3}{c}{\textbf{Fin.}} & \multicolumn{3}{c}{\textbf{H.R.}} & \multicolumn{3}{c}{\textbf{Ind.}} & \multicolumn{3}{c}{\textbf{Phar.}} \\
\midrule
\textbf{Metric} & IoU $\uparrow$ & Recall $\uparrow$ & Time $\downarrow$ & IoU $\uparrow$ & Recall $\uparrow$ & Time $\downarrow$ & IoU $\uparrow$ & Recall $\uparrow$ & Time $\downarrow$ & IoU $\uparrow$ & Recall $\uparrow$ & Time $\downarrow$ & IoU $\uparrow$ & Recall $\uparrow$ & Time $\downarrow$ \\
\midrule
\textbf{\oursys} & 16.0 & 45.7 & 84 & 12.5 & 40.5 & 68 & 12.6 & 47.5 & 15 & 7.3 & 35.6 & 26 & 17.2 & 44.3 & 9 \\
\textbf{w/o Summary} & 15.6 & 45.7 & 194 & 12.6 & 48.7 & 248 & 12.7 & 52.4 & 47 & 7.7 & 39.9 & 73 &  18.2 & 47.6 & 24\\
\textbf{w/o Subnode} & 14.0 & 45.6 & 54 & 10.7 & 44.3 & 32 & 9.9 & 43.3 & 10 & 6.1 & 37.7 & 20 & 16.5 & 45.4 & 5\\
\bottomrule
\end{tabular}%
}
\end{table}

\iffalse
\begin{table}[!t]
\centering
\vspace{-.5em}
\caption{Ablation Study of  Retrieved Chunks on ViDoRe V3 C.S.}
\vspace{-.5em}
\label{tab:ablation3}
\begin{tabular}{l|ccccc}
\toprule
\textbf{Top-k} & 5 & 10 & 20 & 50 & 100 \\
\midrule
\textbf{Acc} & 76.5 & 81.9 & 84.4 & 85.1 & 84.2\\
\textbf{Recall} & 26.0 & 36.0 & 45.7 & 55.3 & 60.8\\
\bottomrule
\end{tabular}%
\vspace{-1em}
\end{table}

\hi{Retrieved Chunks.}
We also test the influence of different retrieved node numbers on C.S. subset of ViDoRe V3. As Table~\ref{tab:ablation3} shows, although a higher number of retrieved nodes brings higher recall, the redundant retrieved information also interfere with model reasoning, resulting in a decline in answer accuracy. And when using LLM API for question answering, more evidence also entails higher invocation costs. Therefore, we ultimately select top 20 as a suitable retrieval hyperparameter.
\fi

\iffalse
\subsection{Limitations}
\label{subsec:limit}
The above experiments and discussions demonstrate that our post-processing is optimized for general scenarios rather than all cases, and may underperform direct text-based or visual RAG when handling queries targeting highly specific content or complex tables in financial reports.

Besides, the automatic annotation by LLMs is not flawless. Although manual observation suggests that the results are generally reliable, the occurrence of some errors cannot be entirely avoided.
\fi

%% file: 5_conclusion.tex
\vspace{-.25cm}
\section{Conclusion}
\label{sec:conclusion}
\vspace{-.25cm}

In this paper, we propose \oursys, a document OCR post-processing pipeline. It employs a fine-tuned model to analyze table truncation, text truncation, title hierarchy, and image-text association based on OCR outputs. Dynamic chunking and synchronization enable it to adapt to long-document processing. Finally, structural and semantic enrichment produces a complete structure that is suitable for downstream tasks like RAG. Our experiments of post-processing on \ourbench, RAG on ViDoRe V3 and end-to-end answering on MMDA confirm the effectiveness of post-processing and its beneficial role in enhancing downstream tasks.

%% file: appendix.tex
\section*{Technical Appendices and Supplementary Material}
\label{sec:appendix}
\section{Dataset Details}
\begin{figure}[htbp]
  \centering

  \begin{minipage}[t]{0.48\linewidth}
    \centering
    \includegraphics[height=0.22\textheight,keepaspectratio]{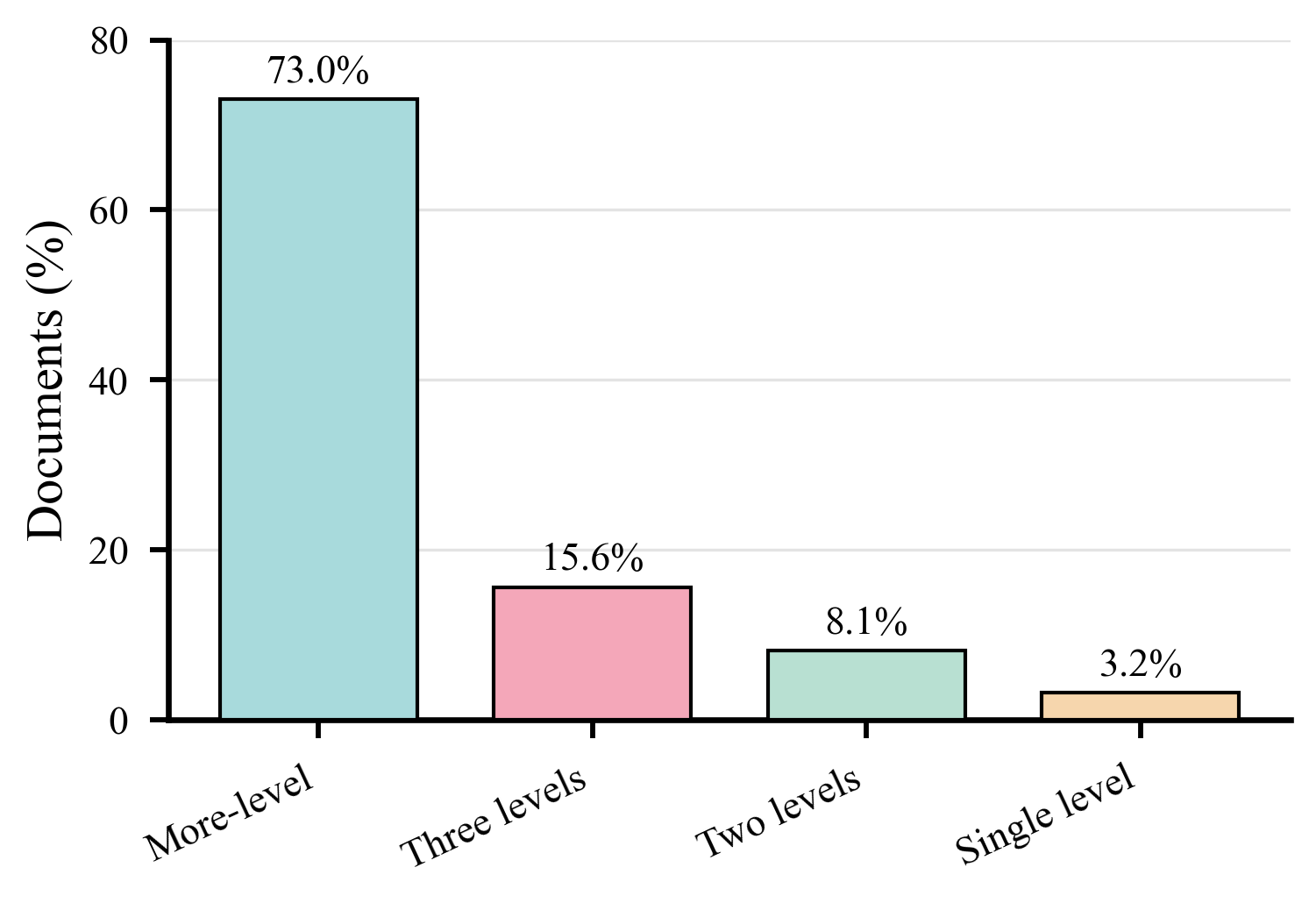}
    \vspace{-1mm}
    \centerline{\small (a) Title hierarchy distribution}
  \end{minipage}
  \hfill
  \begin{minipage}[t]{0.48\linewidth}
    \centering
    \includegraphics[height=0.22\textheight,keepaspectratio]{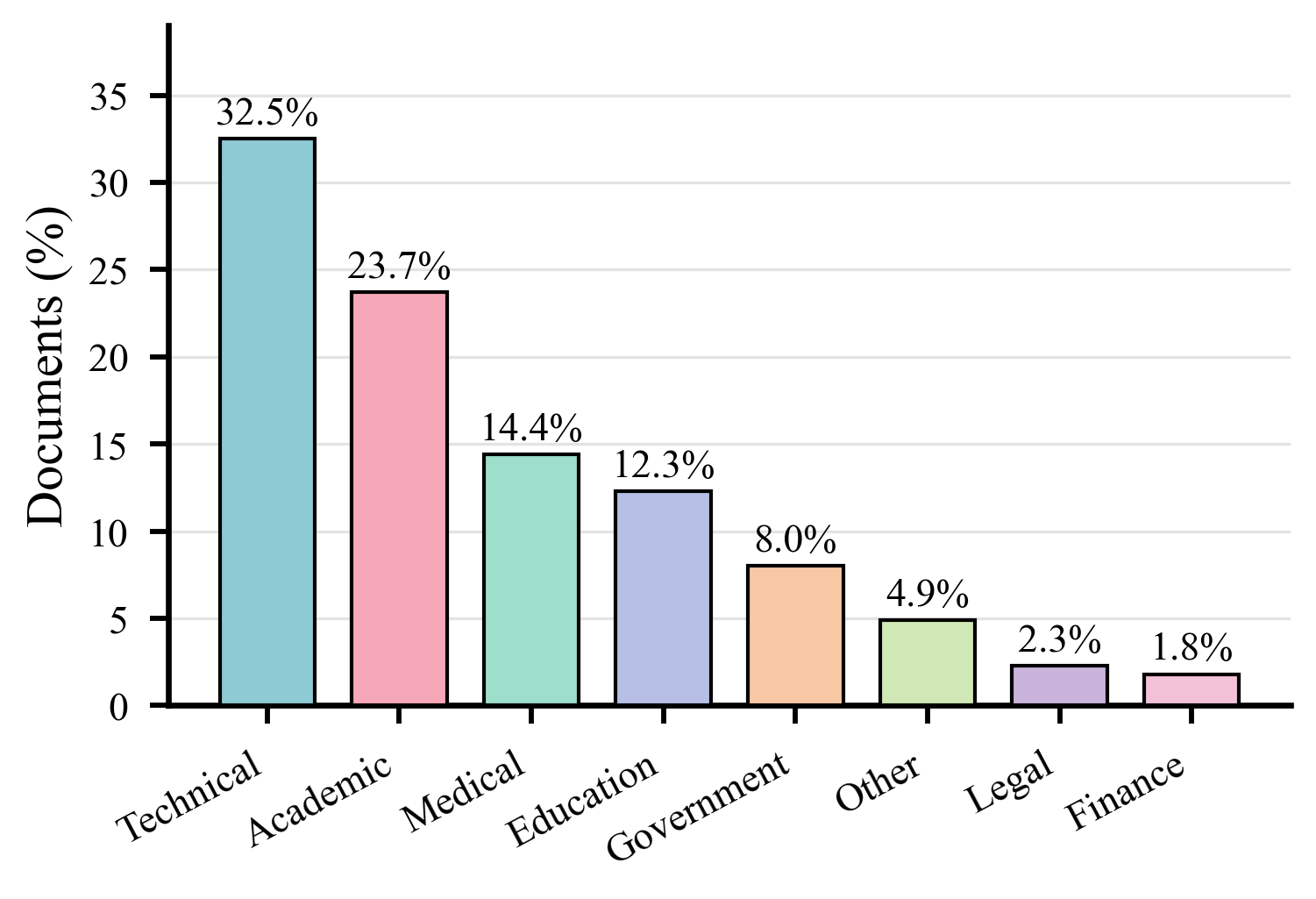}
    \vspace{-1mm}
    \centerline{\small (b) Document domain distribution}
  \end{minipage}

  \caption{Dataset distribution statistics.}
  \label{fig:app-dataset-distribution}
\end{figure}

We construct PostDocBench from a large collection of real-world OCR documents. Since these documents vary substantially in domain, layout, visual style, format, and length, directly sampling from the full collection may lead to biased training data dominated by frequent document types. To obtain diverse and representative PDFs, we first filter valid PDF documents from the returned data pool, and then use a VLM to analyze each document from multiple perspectives, including visual appearance, layout characteristics, document format, content domain, and page length. Based on these analyses, we group documents by domain and length, and sample from different groups to ensure broad coverage of real-world document types. As shown in Fig~\ref{fig:app-dataset-distribution}, this document-level filtering strategy provides a diverse data source and helps improve the generalization ability of the trained post-processing model. Technical reports and academic papers together account for over 50\%, while medical, educational, and governmental documents also account for notable proportions. Financial reports are the least accessible among such public documents due to privacy and copyright concerns.

% 1.我们从大量的回流数据里进行pdf筛选。然后我们利用VLM对大量回流数据进行视觉、版面、格式等风格分析-$>$获取 多样性的pdf 数据来源广泛。 维持训练后的模型的泛化性
% 2. 训练数据如何构建的..
Figure~\ref{fig:app-dataset-distribution} also shows the annotated title hierarchy of these documents. Most documents contain headings with more than three levels. Thus the two-level (doc\_title, para\_title) or three\_level label system used by some OCR models is inadequate.

Based on these documents, we curate the dataset for model training and testing. Specifically, the 165 test instances of PostDocBench encompass 23,178 title elements (title hierarchy), 27,248 pairs (image-text association), 5,761 relation pairs (text truncation), and 1,317 cell-level merge units (table truncation).

% \subsection{Dialect Translation}
% \label{subsec:translation}

\section{Prompts}

We leverage LLMs for annotation, question answering and evaluation with different prompts. Here title hierarchy prompt, text truncation prompt, association prompt and table truncation prompt instruct the LLM to annotate dataset. The input is derived from task-specific filtering of MinerU predictions. QA prompt and evaluation prompt
are used in our experiments of RAG and end-to-end QA.

\begin{promptbox}{Title Hierarchy Prompt}
\promptheading{Instruction}
Given a list of titles (marked by blue boxes in page images) in a document, please analyze title levels according to the page layout, visual clues and semantic infomation.
Identify different levels with different numbers. Use 1 for the highest level title, 2 for the second level, and so on.
Some non-title content may be mixed into the input blocks, and you can mark them with the -1 level.

\promptheading{Few-shot Examples of Title Hierarchy}
\begin{lstlisting}[style=promptstyle]
["2000", "2010", "2020"] ---> ["#2000", "#2010", "#2020"]
["Report", "1.Challenge" , "2.Method", "3.Result"] ---> ["#Report", "##1.Challenge" , "##2.Method", "##3.Result"]
["Schedule", "Day1", "Afternoon", "Night", "Day2", "Morning"] ---> ["#Schedule", "##Day1", "###Afternoon", "###Night", "##Day2", "###Morning"]
\end{lstlisting}

\promptheading{Input Schema}
The input is a \textbf{single JSON object} containing:
\begin{itemize}
\setlength{\itemsep}{0pt}
\setlength{\parskip}{0pt}
\item \promptcode{image}: the image of the document (serving as visual evidence, and multiple pages are separated by black lines)
\item \promptcode{blocks}: a list of title blocks, where each element contains:
  \begin{itemize}
  \setlength{\itemsep}{0pt}
  \setlength{\parskip}{0pt}
  \item \promptcode{idx}: unique identifier of the block (\textbf{must be referenced in the output})
  \item \promptcode{content}: the textual content of the title
  \item \promptcode{page}: number of the page where the block is located
  \item \promptcode{bbox}: location of the block on the page
  \end{itemize}
\end{itemize}

\promptheading{Input Blocks}
\begin{lstlisting}[style=promptstyle]
{{ INPUT_BLOCKS }}
\end{lstlisting}

\promptheading{Output Schema}
\begin{lstlisting}[style=promptstyle]
[
  {
    idx: <int>, # The idx of the title block
    level: <int> # The level of the title (-1 if not a title)
  },
  ...
] 
\end{lstlisting}

\promptheading{Judgment Criteria}
\begin{promptcriteria}
\item A title, semantically, serves as a summary and overarching structure for the main content of a document. Visually, it should stand on its own line and may have a distinct font style.
\item The title numbering in normative documents naturally indicates their hierarchy like 1. | 1.1 | 1.2 | 2.;
\item In page layout, the overall titles of each main block are usually parallel, while multiple titles within a block may be parallel or nested;
\item Visually, multiple consecutive parallel titles usually have the same font and size, and larger font sizes generally indicate higher levels, but visual judgment rules are not always reliable;
\item For title levels cannot be judged by title numbering, layout, and visual features, ultimately determine the hierarchy based on your understanding of the title and context semantics.
\end{promptcriteria}

\promptheading{Note}
Respond strictly in the specified JSON output format. Each input block should be referred once in the output to identify its level.
\end{promptbox}

\begin{promptbox}{Text Truncation Prompt}
\promptheading{Instruction}
Please identify which text blocks in the document are truncated due to column breaks, page breaks, inserted images or tables, and therefore need to be reconnected into a coherent and complete sentence. 
Consider visual factors that may cause truncation, as well as grammatical correctness and semantic coherence from a content perspective. Output the idx of text blocks that you believe require reconnection and briefly explain the reason.

\promptheading{Input Schema}
The input is a \textbf{single JSON object} containing:
\begin{itemize}
\setlength{\itemsep}{0pt}
\setlength{\parskip}{0pt}
\item \promptcode{image}: the image of the document (served as visual evidence, and multiple pages are separated by black lines)
\item \promptcode{blocks}: a list of text blocks with potential truncation, where each element contains:
  \begin{itemize}
  \setlength{\itemsep}{0pt}
  \setlength{\parskip}{0pt}
  \item \promptcode{idx}: unique identifier of the block (\textbf{must be referenced in the output})
  \item \promptcode{content}: completed or broken sentences (the middle parts of long paragraphs are omitted)
  \item \promptcode{page}: number of the page where the block is located
  \item \promptcode{bbox}: location of the block on the page
  \end{itemize}
\end{itemize}

\promptheading{Input Blocks}
\begin{lstlisting}[style=promptstyle]
{{ INPUT_BLOCKS }}
\end{lstlisting}

\promptheading{Output Schema}
\begin{lstlisting}[style=promptstyle]
[
  {
    src: <int>, # The idx of the text block that needs connection at its end 
    tgt: <int>, # The idx of the text block that needs connection at its start
    reason: <str> # Brief explanation. For example, 'The text in block x and block y are broken due to ...; the complete sentence at the connection point is ...'
  },
  ...
] 
\end{lstlisting}

\promptheading{Judgment Criteria}
\begin{promptcriteria}
\item The src and tgt must be two adjacent blocks in Input Blocks.
\item Merge when two text pieces are separated due to layout factors, and the connection can be grammatically and semantically coherent to form a complete sentence.
\item Truncated text blocks usually have similar font style and size.
\item If there is a line break after standalone phrases or complete sentences between src and tgt, no reconnection is needed. 
\end{promptcriteria}

\promptheading{Note}
Respond strictly in the specified JSON output format.
\end{promptbox}

\begin{promptbox}{Association Prompt}
\promptheading{Instruction}
Given a list of element blocks in a document, please analyze the correlation between images, tables and text according to the criteria. 

\promptheading{Input Schema}
The input is a \textbf{single JSON object} containing:
\begin{itemize}
\setlength{\itemsep}{0pt}
\setlength{\parskip}{0pt}
\item \promptcode{image}: the image of the document (served as visual evidence, and multiple pages are separated by black lines)
\item \promptcode{blocks}: a list of titles blocks, where each element contains:
  \begin{itemize}
  \setlength{\itemsep}{0pt}
  \setlength{\parskip}{0pt}
  \item \promptcode{idx}: unique identifier of the block (\textbf{must be referenced in the output})
  \item \promptcode{type}: the type of document elements (related to the criteria)
  \item \promptcode{content}: the textual content of the block (None for images and tables)
  \item \promptcode{page}: number of the page where the block is located
  \item \promptcode{bbox}: location of the block on the page
  \end{itemize}
\end{itemize}

\promptheading{Input Blocks}
\begin{lstlisting}[style=promptstyle]
{{ INPUT_BLOCKS }}
\end{lstlisting}

\promptheading{Output Schema}
\begin{lstlisting}[style=promptstyle]
[
  {
    src: <int>, # The idx of the current block
    tgt: <int>, # The idx of the target block to which the current block should be linked
    reason: <str> # Carefully examine the reasonableness of the correlation to ensure it aligns with the actual document structure and the criterias
  },
  ...
] 
\end{lstlisting}

\promptheading{Judgment Criteria}
\begin{promptcriteria}
\item The block with type \promptcode{image} or \promptcode{table} should be linked to the most related \promptcode{title} block;
\item The block with type \promptcode{image\_caption} or \promptcode{image\_footnote} should be linked to the most related \promptcode{image} block;
\item The block with type \promptcode{table\_caption} or \promptcode{table\_footnote} should be linked to the most related \promptcode{table} block;
\item Links not falling into the above three situations cannot be connected.
\end{promptcriteria}

\promptheading{Note}
Respond strictly in the specified JSON output format. You can only output links that meet the criterias.
\end{promptbox}

\begin{promptbox}{Table Truncation Prompt}
\promptheading{Instruction}
Given two table fragments split across consecutive PDF pages, determine whether they belong to the same cross-page table and identify which columns require semantic merging.

\promptheading{Input Schema}
The input contains:
\begin{itemize}
\setlength{\itemsep}{0pt}
\setlength{\parskip}{0pt}
\item \promptcode{upper\_caption}: the caption of the table on the previous page
\item \promptcode{upper\_row}: the last few rows of the table on the previous page
\item \promptcode{lower\_caption}: the caption of the table on the next page
\item \promptcode{lower\_row}: the first few data rows of the table on the next page
\end{itemize}

\promptheading{Input Blocks}
\begin{lstlisting}[style=promptstyle]
{{ INPUT_BLOCKS }}
\end{lstlisting}

\promptheading{Output Schema}
\begin{lstlisting}[style=promptstyle]
[
    {
        judgement:<list> # A list indicating cell-level merging decisions
    }
]
\end{lstlisting}

\promptheading{Judgment Criteria}
\begin{promptcriteria}
\item If the two table fragments have different column counts or clearly cannot belong to the same table, output [].
\item Otherwise, output exactly one object with a judgment list.
\item The length of judgment must be equal to the number of columns in the table. Each value corresponds to one column from left to right.
\item Use 1 if the two cells in the same column should be semantically merged into one logical cell, such as in cases of hyphenation, an incomplete phrase, a split date/number, or rowspan continuation.
\item Use 0 if the two cells in the same column are independent row contents, even if they are similar or identical.
\item For empty cells, make the decision based on the overall row context and table captions.
\end{promptcriteria}

\promptheading{NOTE}
Respond strictly in the specified JSON output format. Do not explain your reasoning.
\end{promptbox}

\begin{promptbox}{QA Prompt}

You are an expert at answering queries based on documents.

Here is a list of relevant document contents: \promptcode{\{data\}}

Based on the above contents, answer the following query: \promptcode{\{query\}}

Keep the response short when appropriate. Output the answer only.

\end{promptbox}

\begin{promptbox}{Evaluation Prompt}
You are an expert judge evaluating the accuracy of a test answer against a gold-standard true answer.
Your goal is to determine if the test answer captures the essential \promptcode{"core information."}

\promptheading{Evaluation Criteria:}
\begin{itemize}
\setlength{\itemsep}{0pt}
\setlength{\parskip}{0pt}
\item Correct: The test answer contains all core information of the true answer. Minor omissions of non-essential details or the addition of minor, non-contradictory information should still be marked as \promptcode{"Correct."}
\item Partially Correct: The test answer captures some of the core information, but suffers from significant omissions or includes substantial extra information that was not requested or present in the true answer.
\item Incorrect: The test answer is fundamentally wrong, contradicts the true answer, or misses the core information entirely.
\end{itemize}

\promptheading{Input Data:}
\begin{lstlisting}[style=promptstyle]
Query: {query}
True Answer: {true_answer}
Test Answer: {test_answer}
\end{lstlisting}

\promptheading{Output Format:}
Provide a very brief explanation for your judgment. You must output your final response in JSON format with two fields: \promptcode{"explanation"} and \promptcode{"judgment"} (which must be \promptcode{"Correct"}, \promptcode{"Partially Correct"}, or \promptcode{"Incorrect"}).
\end{promptbox}
\section{Cases}

\begin{figure}[h]
    \centering
    \includegraphics[width=\linewidth]{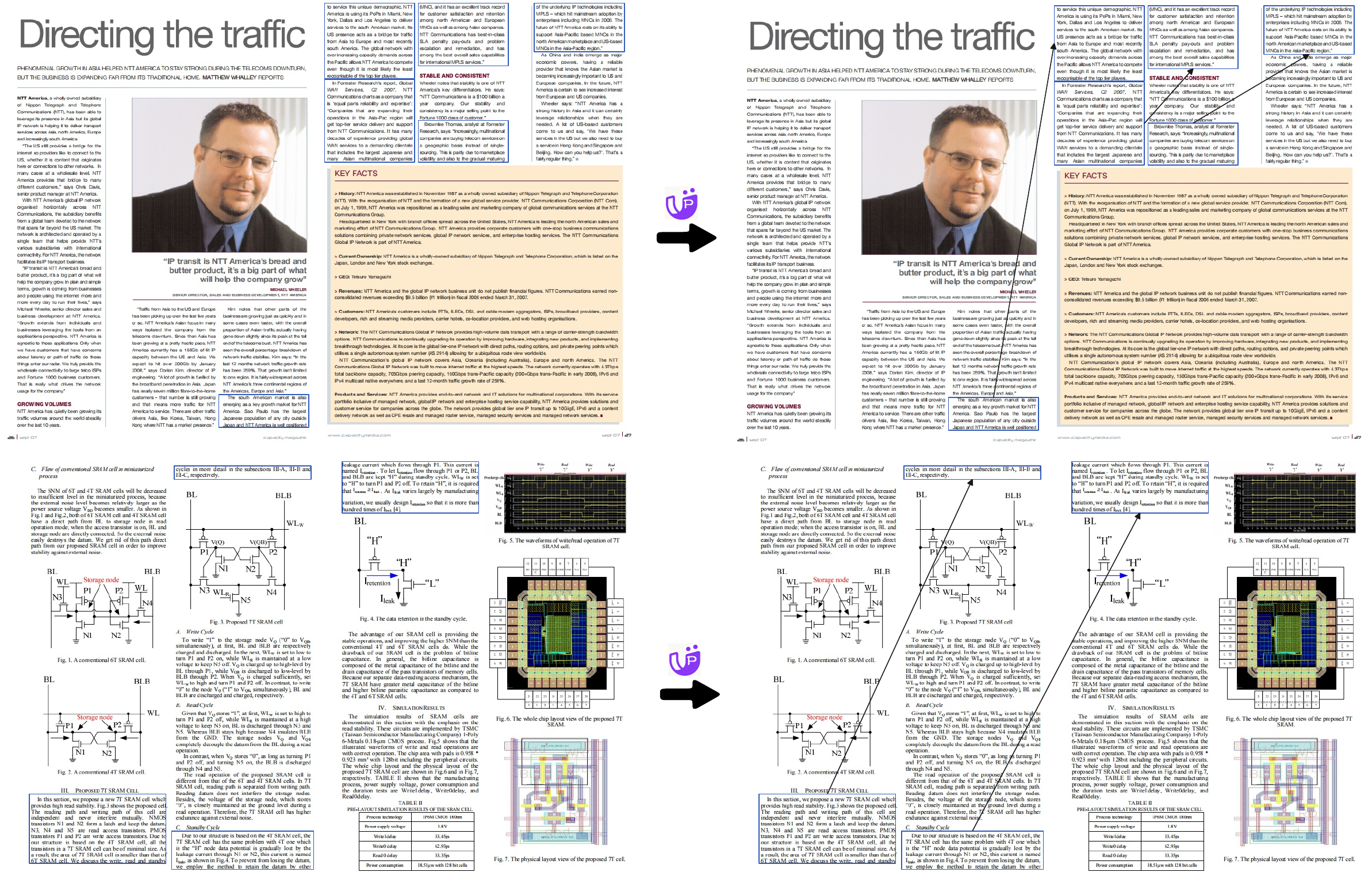}
    \vspace{-1.75em}
    \caption{Cases of Text Truncation Analysis.}
    \vspace{-0.8em}
    \label{fig:textmerge_cases}
\end{figure}

\begin{figure}[h]
    \centering
    \includegraphics[width=\linewidth]{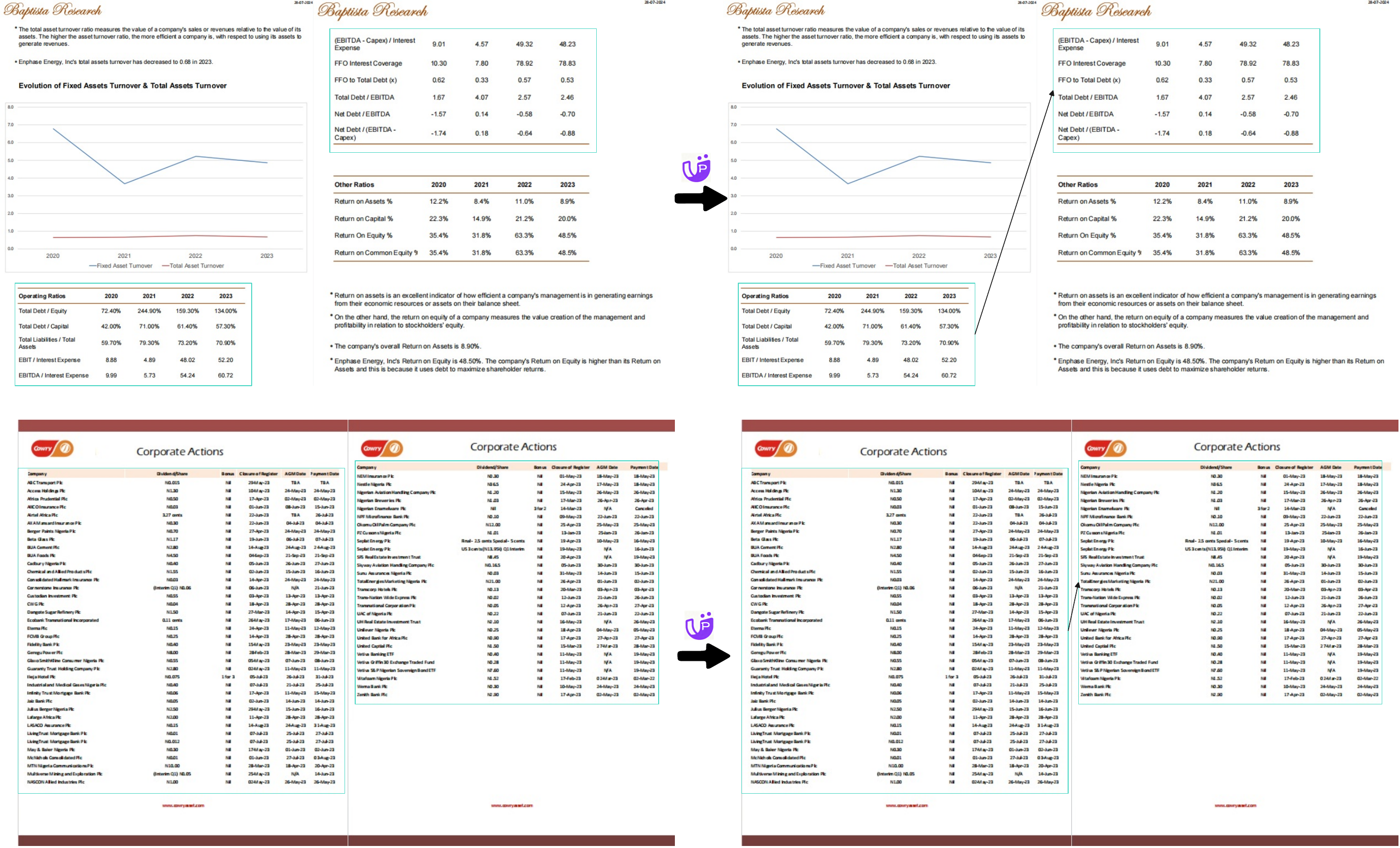}
    \vspace{-1.75em}
    \caption{Cases of Table Truncation Analysis.}
    \vspace{-0.8em}
    \label{fig:tabmerge_cases}
\end{figure}

\begin{figure}[h]
    \centering
    \includegraphics[width=\linewidth]{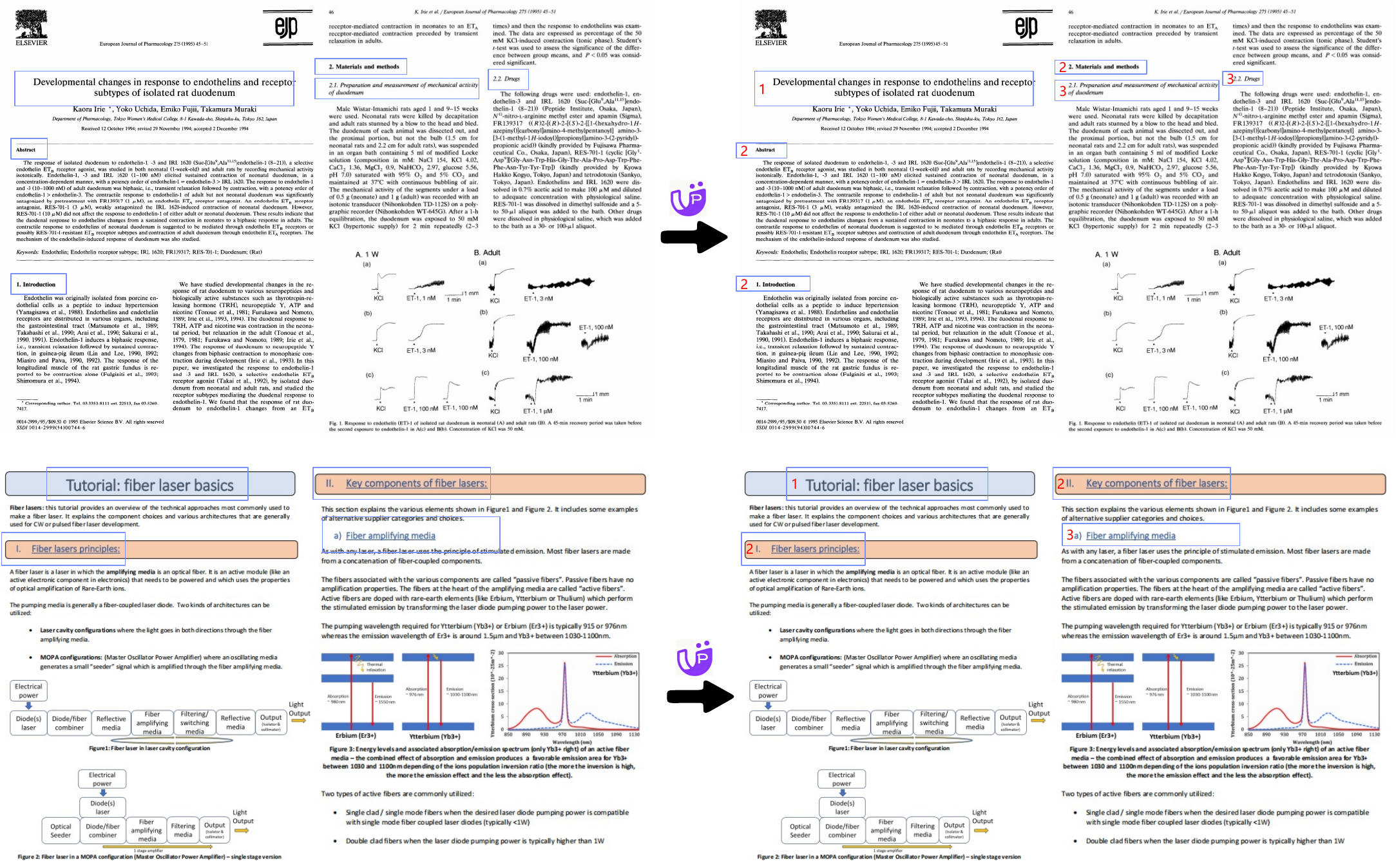}
    \vspace{-1.75em}
    \caption{Cases of Title Hierarchy Analysis.}
    \vspace{-0.8em}
    \label{fig:title_cases}
\end{figure}

\begin{figure}[h]
    \centering
    \includegraphics[width=\linewidth]{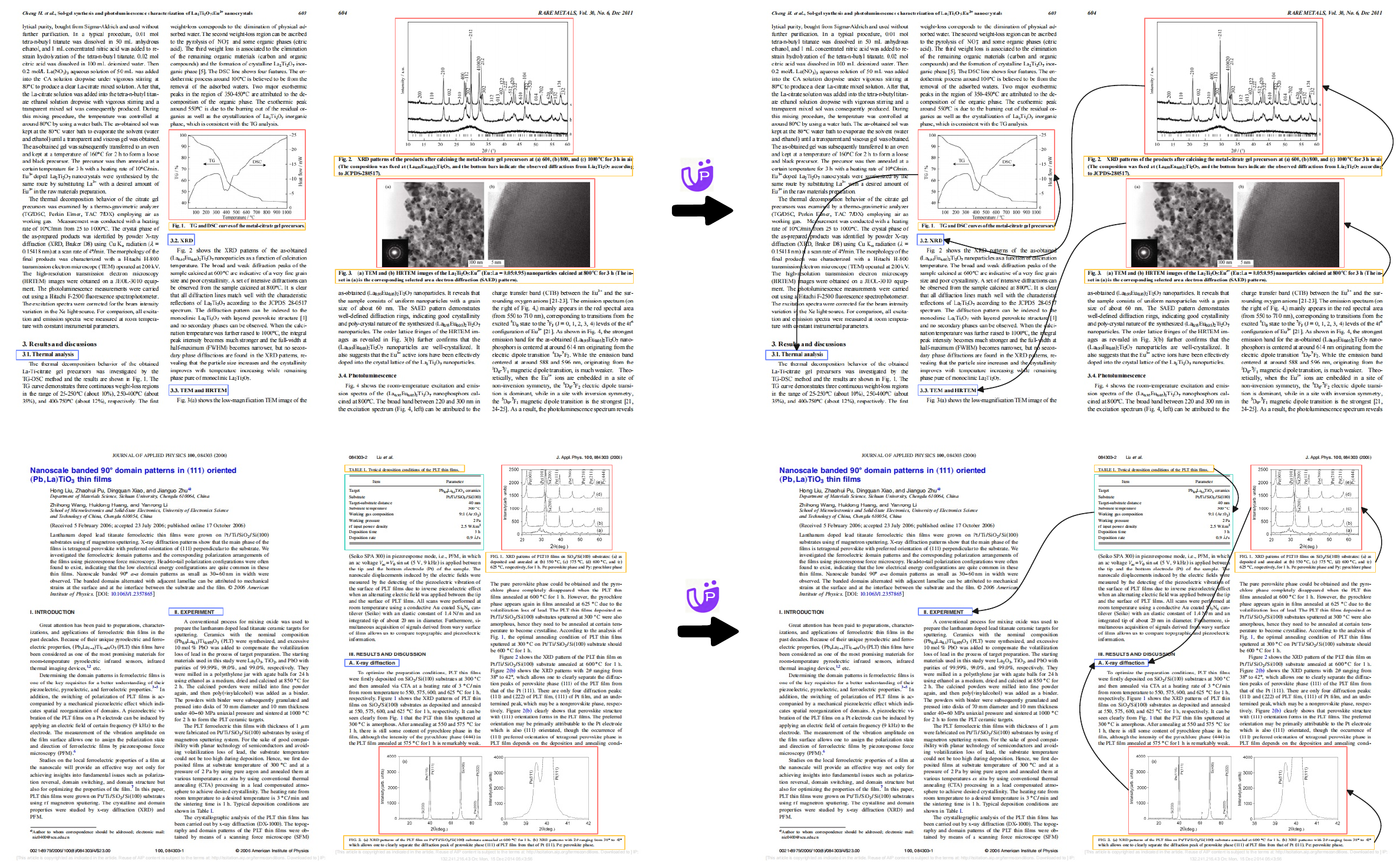}
    \vspace{-1.75em}
    \caption{Cases of Text-Image Association Analysis.}
    \vspace{-0.8em}
    \label{fig:association_cases}
\end{figure}